\documentclass[journal]{IEEEtran}

\usepackage{amssymb,amsmath,amsfonts}
\usepackage{algorithmic}
\usepackage{algorithm}
\usepackage{array}
\usepackage{bbold}
\usepackage{booktabs}
\usepackage{colortbl}
\usepackage[hidelinks]{hyperref}
\usepackage{graphicx}
\usepackage{makecell}
\usepackage[mathlines, switch]{lineno}
\usepackage{mathtools}
\usepackage{multirow}
\usepackage[sort&compress, numbers]{natbib}
\usepackage{pifont}
\usepackage{siunitx}
\usepackage{stfloats}
\usepackage[caption=false,font=footnotesize]{subfig}
\usepackage{textcomp}
\usepackage{xurl}
\usepackage{verbatim}
\usepackage[dvipsnames]{xcolor}

\hypersetup{
  colorlinks   = true, %Colours links instead of ugly boxes
  urlcolor     = CadetBlue, %Colour for external hyperlinks
  linkcolor    = Cerulean, %Colour of internal links
  citecolor   = Maroon %Colour of citations
}

\begin{document}

\title{Choose Your Game Wisely: Measuring Game-Theoretic Structures in Real-World Vehicle Interactions}

\author{Yueyuan~Li, Rongcheng~Nie, Weijie~Xi, Mingyang~Jiang, Songan~Zhang,  Hanyang~Zhuang, and~Ming~Yang % <-this % stops a space
\thanks{This work is supported in part by the National Natural Science Foundation of China under Grant 62573289, 52402504, and U22A20100 \textit{(Corresponding author: Ming Yang, email: MingYANG@sjtu.edu.cn)}.}% <-this % stops a space
\thanks{Yueyuan~Li, Weijie~Xi, Mingyang~Jiang, Ming~Yang are with the School of Automation and Intelligent Sensing, Shanghai Jiao Tong University, Key Laboratory of System Control and Information Processing, Ministry of Education of China, Shanghai, 200240, CN.}
\thanks{Rongcheng Nie is with the School of Airspace Science and Engineering, Shandong University, Weihai, Shandong, 264209, CN.}
\thanks{Songan~Zhang is with the Global Institute of Future Technology, Shanghai Jiao Tong University, Shanghai, 200240, CN.}
\thanks{Hanyang~Zhuang is with Global College, Shanghai Jiao Tong University, Shanghai, 200240, CN.}
}

\maketitle

% !TEX root = ../main.tex

\begin{abstract}
Game-theoretic models provide principled frameworks for modeling vehicle interactions, but their underlying temporal assumptions have not been systematically examined against real-world driving behavior.
In particular, it remains unclear how simultaneous, sequential, and asymmetric interaction structures can be measured from vehicle trajectories.
This paper develops a trajectory-based interaction measurement framework to identify interaction events and quantify behavioral change onset, temporal organization, post-onset response dynamics, and ordering stability.
The framework uses behavioral deviations to verify candidate interactions.
We evaluate the framework on six real-world trajectory datasets, including INTERACTION, highD, inD, rounD, Waymo Open Motion, and nuPlan, covering diverse road geometries, traffic environments, and interaction types.
The results show that concurrent and sequential behavioral changes both constitute substantial proportions of observed following, merging, and conflicting interactions.
Among sequential interactions, stable ordering is more prevalent than alternating ordering, indicating that persistent asymmetric roles are a common interaction structure.
Importantly, temporal precedence does not necessarily coincide with a measurable behavioral response, indicating that temporal ordering alone may not be sufficient to characterize behavioral dependence.
These findings show that real-world interactions exhibit concurrent, sequential, and persistently ordered temporal structures.
Different game-theoretic formulations are therefore better regarded as complementary modeling abstractions for different interaction regimes rather than as a universal structure governing all vehicle interactions.
\end{abstract}

% !TEX root = ../main.tex

\section{Introduction}

Vehicle behavior in complex traffic is often shaped by surrounding vehicles rather than determined independently~\cite{wang2022social}.
Such interactions are particularly evident in highways, ramps, intersections, and roundabouts~\cite{shirazi2016looking,zhang2024identifying}, where a behavioral change by one vehicle may induce a subsequent response from another.
Real-world trajectory data provide direct measurements of vehicle motion and have been widely used to identify interacting vehicles~\cite{jiang2025naturalistic}, characterize interaction patterns~\cite{zhang2024identifying,abarghooei2026driving}, and model or predict vehicle behavior~\cite{liang2025interaction}.
However, existing studies primarily detect or model observed behavior, while the temporal and directional organization of behavioral responses remains less systematically measured.
In particular, proximity or temporal precedence alone does not establish whether an observed change constitutes a measurable behavioral response to another vehicle.

Game theory provides explicit models of interactive decision making by specifying how interacting vehicles make decisions.
Both cooperative and non-cooperative formulations have been considered in driving applications~\cite{hang2021cooperative,liniger2019noncooperative}, with Nash and Stackelberg formulations among the commonly used approaches~\cite{talebpour2015modeling,wang2015game}.
Nash formulations typically assume simultaneous decisions, whereas Stackelberg formulations impose asymmetric leader--follower roles~\cite{nash1950equilibrium,von2010market}.
These formulations therefore imply distinct temporal and directional structures in observable behavior.
Whether real-world interactions exhibit these structures, however, has not been systematically measured.

To address this gap, we develop a trajectory-based interaction measurement framework that identifies interaction events and quantifies behavioral change onset, temporal organization, response dynamics, and ordering stability.
Interaction directionality captures which vehicle changes its behavior first and whether a subsequent measurable response occurs, while temporal organization and ordering stability characterize the timing and persistence of these roles.
We then use these measurements to assess which game-theoretic temporal structures best characterize different observed interaction regimes.

Rather than proposing another interaction model, this study establishes an empirical basis for choosing between alternative temporal interaction structures when modeling real-world driving.
The contributions of this paper are as follows:
\begin{itemize}

    \item An interaction measurement framework that separates behavioral onset, temporal organization, response dynamics, and ordering stability, enabling interaction structures to be quantified directly from real-world trajectories.

    \item A cross-dataset characterization showing that concurrent and sequential behavioral changes coexist across following, merging, and conflicting interactions, while sequential interactions are predominantly characterized by stable rather than alternating ordering.

    \item An empirical assessment showing that concurrent, sequential, and persistently ordered interaction structures coexist in real-world driving, supporting a complementary view of simultaneous-move, sequential-move, and leader--follower formulations across interaction regimes.

\end{itemize}

% !TEX root = ../main.tex

\section{Related Works}

\subsection{Interaction Behavior Analysis}

The availability of large-scale real-world trajectory datasets has enabled vehicle interactions to be studied directly from observed driving behavior.
Datasets such as INTERACTION, nuPlan, and the Waymo Open Motion Dataset (WOMD) contain numerous interactive traffic scenarios and have been widely used for interaction analysis and motion prediction~\cite{zhan2019interaction,caesar2021nuplan,ettinger2021large}.
Vehicle interaction has also been studied from behavioral, cognitive, optimization-based, learning-based, and game-theoretic perspectives~\cite{wang2022social,crosato2024social}.
For this study, trajectory-based analysis is especially relevant because it provides direct observations of how interacting vehicles adjust their behavior over time.

A number of studies have used trajectory data to identify interaction patterns.
Zhang and Wang~\cite{zhang2019learning} extracted vehicle-to-vehicle interaction patterns at signalized intersections using driving primitives and unsupervised clustering.
Zhang et al.~\cite{zhang2024identifying} considered how interaction patterns evolve during mandatory and discretionary lane changes using graph structures and hidden Markov models.
Jiang et al.~\cite{jiang2025naturalistic} developed InterHub to extract and organize dense interaction events from large-scale driving datasets.
Other work has focused more directly on identifying which surrounding vehicles are behaviorally relevant.
Hazard et al.~\cite{hazard2022importance} ranked surrounding agents according to their effect on the ego vehicle's planned trajectory, while Wang et al.~\cite{wang2026quantitative} introduced a Vehicle Interaction Level to quantify how strongly the behavior of one vehicle is constrained by another.
These studies move beyond detecting interaction events and provide measures of interaction relevance or strength.

Interaction has also been modeled explicitly in trajectory prediction.
Some early methods incorporated semantic interaction features or graph-based relations between surrounding vehicles~\cite{hong2019rules,li2019grip}.
Lee et al.~\cite{lee2019joint} jointly predicted trajectories and pairwise interaction modes, and Kumar et al.~\cite{kumar2021interaction} represented temporal interaction types as edges in a hybrid traffic graph.
M2I~\cite{sun2022m2i} modeled directed dependence by dividing interacting agents into influencer--reactor pairs and conditioning the reactor's prediction on the influencer.
Recent work continues to introduce more explicit forms of vehicle dependence into motion prediction~\cite{liang2025interaction}.

Overall, trajectory-based studies have become increasingly detailed in describing whether vehicles interact, which vehicles are involved, and how strongly their behaviors are related.
In most cases, however, these descriptions are introduced to support interaction identification or trajectory prediction.
How the behavioral dependence between two vehicles develops within an interaction episode is less often examined directly from observed trajectories.

\subsection{Interaction Modeling Based on Game Theory}

Game theory has been widely used to model interactive driving because it explicitly accounts for the interdependence between vehicle decisions.
In a game-theoretic formulation, each vehicle is typically associated with an objective that reflects factors such as safety, efficiency, and comfort~\cite{wang2015game}.
The objective may also incorporate the outcomes of other road users, allowing different social preferences, ranging from predominantly self-interested behavior to more cooperative behavior~\cite{schwarting2019social}.

Game-theoretic models differ not only in vehicle objectives but also in how the interaction between vehicles is structured.
Nash formulations generally model mutually dependent decisions within the same decision stage~\cite{wang2015game, zhu2023sequential}.
Stackelberg formulations instead introduce an asymmetric and sequential relationship, in which one vehicle is treated as the leader and another responds as the follower~\cite{yoo2013stackelberg, hang2020human}.
Hierarchical formulations introduce additional reasoning levels or decision stages to describe more involved interactions~\cite{fisac2019hierarchical, huang2024game}.
Nash and Stackelberg formulations have also been considered within the same driving decision framework to examine how different game structures affect decision making~\cite{hang2020human, hang2022driving}.

Some game-based methods allow parts of the interaction to change as more observations become available.
Zhang et al.~\cite{zhang2019game}, for example, estimated driver aggressiveness online within a Stackelberg-based controller for mandatory lane changing.
Li et al.~\cite{li2019decision} used Bayesian inference together with cognitive hierarchy theory to update the ego vehicle's strategy according to the inferred behavior of another driver.
Liu et al.~\cite{liu2022interaction} treated pairwise leader--follower relationships as uncertain in a partially observable game and estimated them from observed trajectories during forced merging.
In these models, driver type, intention, or role does not have to remain fixed throughout the interaction, although such adaptation is still performed within a specified game formulation.

These formulations have been applied to a variety of driving scenarios, including lane changing, merging, intersections, roundabouts, and racing~\cite{lin2025scenario}.
Although the specific assumptions differ, the interaction structure is generally specified as part of the model formulation~\cite{di2020liability}, while real-world or simulated driving data are mainly used to calibrate model parameters, reproduce observed behavior, or evaluate model performance~\cite{yoo2013stackelberg,hang2020human}.
Consequently, empirical evidence is often used to assess behavior generated under a predefined interaction structure rather than to identify that structure directly from observed interactions.

Taken together, existing studies either represent interaction for behavioral analysis and prediction or prescribe its structure within a behavioral model.
This leaves the structure observed in real driving behavior less well understood.

% !TEX root = ../main.tex

\section{Method}

\subsection{Preliminary}\label{sec:preliminary}

Before introducing the interaction measurement framework, we review the observable temporal implications of assumptions underlying commonly used game-theoretic models for vehicle interaction.

\subsubsection{Simultaneous-move assumption}

In simultaneous-move games, players select their actions within the same decision stage without observing the current actions of other players.
The Nash equilibrium provides a classical solution concept for such games, where each player's strategy is optimal given the strategies selected by others~\cite{nash1950equilibrium}.
In vehicle interactions, this corresponds to concurrent behavioral changes without a measurable temporal precedence between the vehicles.

\subsubsection{Sequential-move assumption}

Sequential-move games describe interactions in which players make decisions at different stages, creating an ordered decision process where the action of one player precedes and can be observed before the subsequent decision of another player~\cite{fudenberg1991game}.
In vehicle interactions, this corresponds to a temporally ordered sequence of observable behavioral changes.

\subsubsection{Leader--follower structure}

Stackelberg games represent a sequential interaction with asymmetric leader--follower roles, where the follower responds after observing the leader's action~\cite{yoo2013stackelberg}.
In vehicle interactions, this structure implies persistent asymmetric ordering, which can be empirically examined from observed behavioral changes.

\subsection{Interaction Event Extraction}\label{sec:interaction-event-extraction}

Following the concept of interaction in~\cite{markkula2020defining}, we consider an interaction as a situation in which two vehicles intend to occupy the same spatial region at the same time in the near future.
Such interactions may be expressed through changes in vehicle motion as well as non-kinematic cues such as vehicle lights, gaze, or gestures.
In this study, we focus on interactions observable from vehicle trajectories.

\begin{figure*}[htb]
    \centering
    \subfloat[Following]{\includegraphics[width=0.24\linewidth]{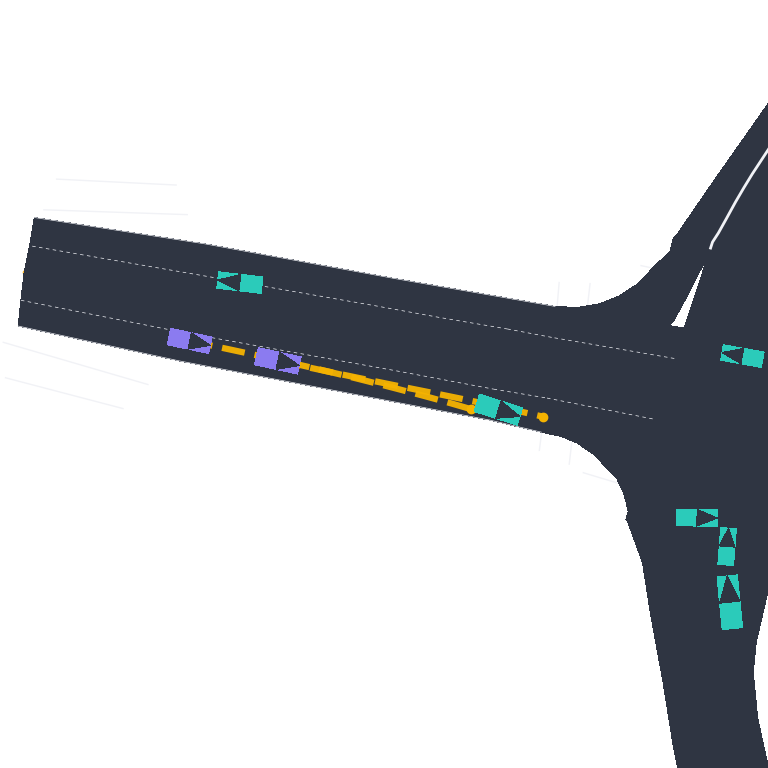}}
    \hfill
    \subfloat[Merging]{\includegraphics[width=0.24\linewidth]{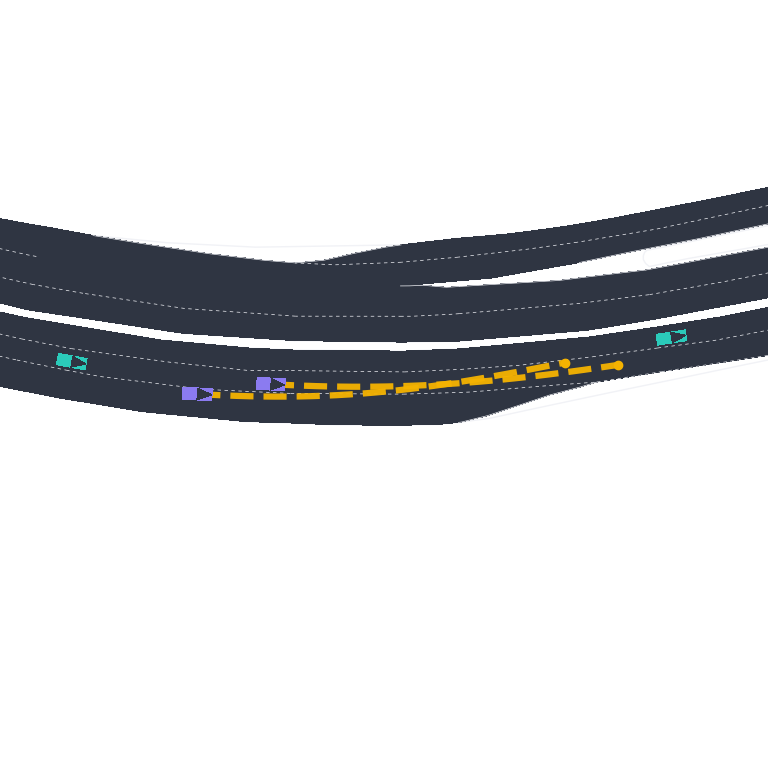}}
    \hfill
    \subfloat[Conflicting: intersecting paths]{\includegraphics[width=0.24\linewidth]{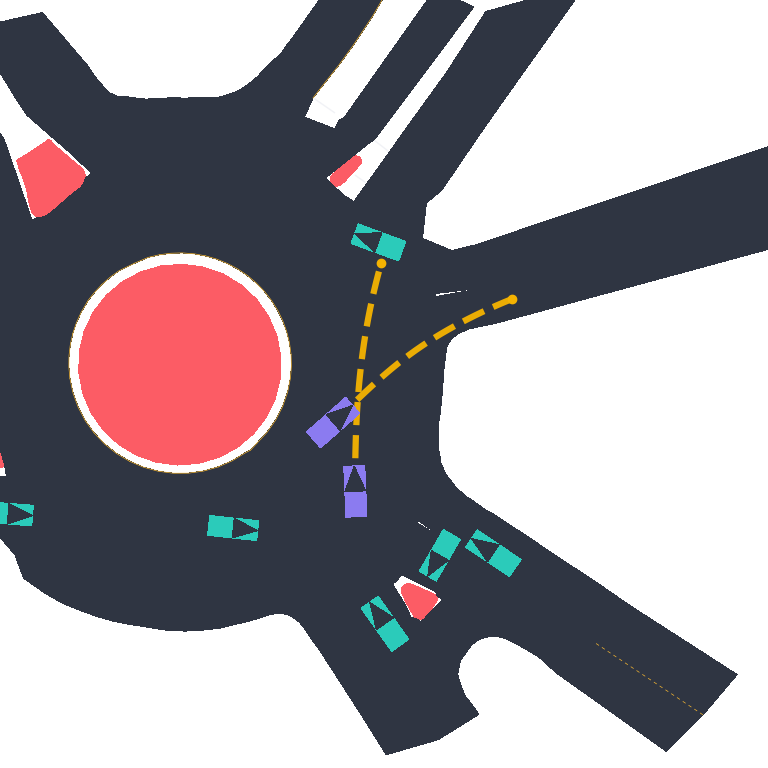}}
    \hfill
    \subfloat[Conflicting: opposing paths]{\includegraphics[width=0.24\linewidth]{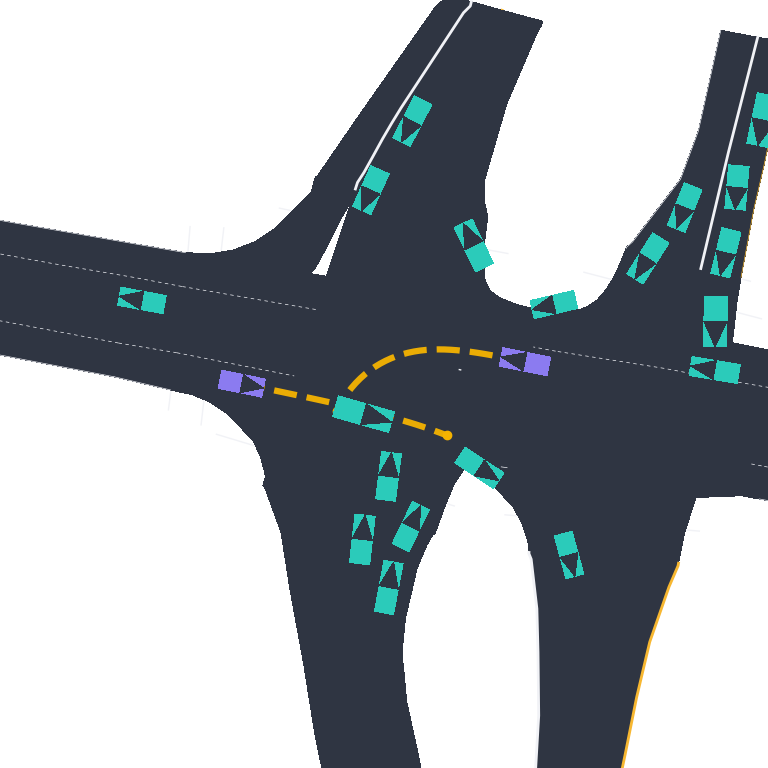}}
    \caption{Examples of the three interaction types: following, merging, and conflicting, from the INTERACTION dataset~\cite{zhan2019interaction}.}\label{fig:typical-interaction}
\end{figure*}

We categorize trajectory-level relations into three interaction types: following, merging, and conflicting, consistent with common interaction patterns considered in previous studies~\cite{wang2022social,zhang2019learning,zhang2024identifying}.
\textbf{Following (F)} describes vehicles traveling in the same direction on the same lane.
\textbf{Merging (M)} describes vehicles traveling in the same direction on different lanes, with one vehicle moving toward the other vehicle's lane.
\textbf{Conflicting (C)} describes vehicles whose future paths intersect or converge from different directions, including both intersecting and opposing-path configurations.
Figure~\ref{fig:typical-interaction} illustrates representative examples of these interaction types in different traffic scenarios, including highways, urban roads, and roundabouts.

For potential interaction detection, each vehicle is assigned a reference trajectory that maintains its current speed and follows its lane route toward the observed destination.
The observed and reference trajectories are converted into vehicle-occupancy-aware motion corridors.
Potential following, merging, and conflicting interactions are detected from road topology, spatial compatibility, and temporal compatibility~\cite{zhang2019learning,li2019grip,kumar2021interaction,sun2022m2i,zhang2024identifying,wang2026quantitative}, and frame-level detections are aggregated into candidate interaction events.

Spatiotemporal compatibility alone, however, does not establish a behavioral response.
We therefore use speed deviation as additional behavioral evidence.
For each candidate event window $W$, we construct a constant-velocity reference by holding vehicle $i$'s speed at the beginning of $W$ while following its lane route, denoted by $v_i^{w}(t)$, and quantify the deviation of the observed speed $v_i^{\mathrm{obs}}(t)$ from this reference within the window.
Because vehicles differ in typical speed variability, we normalize this deviation against a vehicle-specific empirical baseline: for control windows in which vehicle $i$ is not involved in any candidate relation, we compute the same deviation and take its mean $\mu_i$ and standard deviation $\sigma_i$.
The normalized deviation is
\begin{equation}
z_i
=
\frac{
\operatorname*{median}_{t\in W}
\left|
v_i^{\mathrm{obs}}(t)-v_i^{w}(t)
\right|-\mu_i
}
{\sigma_i+\epsilon_b},
\label{eq:normalized-deviation}
\end{equation}
where $\epsilon_b = 0.5$~m/s is a stabilization term.
A candidate pair $(i,j)$ is retained if at least one participant exhibits a sufficiently large normalized deviation:
\begin{equation}
\max(z_i,z_j)>\tau,
\label{eq:behavior-confirmation}
\end{equation}
where $\tau=2.0$ is the deviation threshold.
This criterion retains asymmetric cases in which only one participant exhibits a marked behavioral adjustment.

For merging and conflicting candidates, we require the vehicles to encounter the shared spatial region within a temporal gap of at most 1.0 s, preventing temporally unrelated behaviors from being associated with the same event.

\subsection{Temporal Characterization of Interaction}\label{sec:temporal-characterization}

\subsubsection{Behavioral Change Onset}

We first identify behavioral change onset relative to a map-constrained reference trajectory.
This reference trajectory assumes that the vehicle follows its lane route toward the observed destination while accounting for road geometry and planned deceleration, such as curvature and stop-line constraints.
For vehicle $i$, the speed residual at time $t$ is defined as
\begin{equation}\label{eq:speed-residual}
    r_i(t) = v_i^{\mathrm{obs}}(t) - v_i^{\mathrm{ref}}(t),
\end{equation}
where $v_i^{\mathrm{obs}}(t)$ and $v_i^{\mathrm{ref}}(t)$ denote the observed and map-constrained reference speeds, respectively.
To suppress trajectory noise, a behavioral change must exceed a residual threshold continuously for a minimum duration.
The behavioral change onset is defined as
\begin{equation}\label{eq:behavioral-change}
    t_i^{\mathrm{onset}}
    =
    \inf
    \left\{
    t:
    |r_i(t')| > \delta,\;
    \forall t' \in [t,t+d_{\min})
    \right\},
\end{equation}
where the residual threshold is set to $\delta=0.3$ m/s and the minimum persistence duration is set to $d_{\min}=400$ ms.

\subsubsection{Temporal Organization}

We use the relative timing of behavioral change onsets to characterize whether the two participants exhibit temporally aligned or temporally ordered behavioral changes.
For an interaction event involving vehicles $i$ and $j$, the onset difference is defined as
\begin{equation}\label{eq:onset-difference}
    \Delta t_{ij}
    =
    t_i^{\mathrm{onset}}
    -
    t_j^{\mathrm{onset}}.
\end{equation}
When both onsets are detected, an event is classified as \textbf{concurrent} if $|\Delta t_{ij}|\leq\tau_{\mathrm{tol}}$ and as \textbf{sequential} otherwise, where $\tau_{\mathrm{tol}}=400$ ms.
Thus, concurrent denotes no measurable temporal precedence at this tolerance.
For sequential events, the sign of $\Delta t_{ij}$ indicates which vehicle exhibits an observable behavioral change first.
An event is classified as \textbf{one-sided} when only one onset is detected and as \textbf{unresolved} when neither onset is detected.
All proportions use the full set of verified interaction events; one-sided and unresolved events are therefore counted as non-concurrent.

\subsubsection{Role Stability}

The first-onset comparison captures only the initial temporal relationship between the two vehicles.
We therefore examine the complete sequence of detected behavioral changes within each interaction.

Events whose detected changes fall within the temporal tolerance are classified as \textbf{concurrent}; otherwise, changes are ordered chronologically according to the corresponding vehicle.
Events are classified as \textbf{stable-ordering} when fewer than two ordering reversals occur, and as \textbf{alternating} when two or more reversals are observed.
Events with changes detected for only one vehicle are classified as \textbf{one-sided}, while events without detectable changes are classified as \textbf{unresolved}.
The sequence-level concurrent class is distinct from the first-onset classification above; all temporal-organization statistics and the GLMM use the sequence-level classification.

\subsubsection{Post-onset Response Timing}

For each directed pair $i\rightarrow j$, we examine whether vehicle $j$ exhibits a new behavioral change after the onset of vehicle $i$.
The response threshold for the target vehicle is defined as
\begin{equation}
    \theta_j
    =
    \max
    \left(
    \delta,\,
    \mu_{j,\mathrm{pre}}
    +
    k\sigma_{j,\mathrm{pre}}
    \right),
\end{equation}
where $\mu_{j,\mathrm{pre}}$ and $\sigma_{j,\mathrm{pre}}$ are the mean and standard deviation of $|r_j(t)|$ before the source change, respectively, and $k=2$.

Each directed pair is classified as \textbf{response}, \textbf{preactive}, \textbf{no-response}, or \textbf{right-censored}.
Directions without a valid source onset or with fewer than three pre-onset baseline frames are recorded separately.
A response is detected when the target exhibits a sustained residual above $\theta_j$ after the source change, whereas a preactive target has already exhibited behavioral change before the source onset.

Let $t_j^{\mathrm{response}}$ denote the first sustained post-onset change of vehicle $j$ satisfying the response criterion.
For response directions, the response lag is
\begin{equation}
    \Delta t_{i\rightarrow j}
    =
    t_j^{\mathrm{response}}
    -
    t_i^{\mathrm{onset}}.
\end{equation}
Response-lag statistics are computed only over directions classified as response.
The post-onset search window is limited to 2000~ms; directions without a qualifying response within this window are treated according to the corresponding no-response or right-censored criteria.

\subsection{Statistical Analysis}

\subsubsection{Cluster-aware Uncertainty Estimation}

Confidence intervals are estimated using scene-level cluster bootstrap resampling.
For each bootstrap replicate, scenes are sampled with replacement and all interaction events belonging to the sampled scenes are retained.
We use 2,000 bootstrap replicates and report percentile-based 95\% confidence intervals.

\subsubsection{Interaction-type Heterogeneity}

To compare temporal organization across interaction types while accounting for event- and scene-level heterogeneity, we fit a logistic mixed-effects model of concurrent organization on interaction type, event duration, and number of vehicles, with a scene-level random intercept.
Interaction-type effects are reported as adjusted odds ratios with 95\% confidence intervals.

% !TEX root = ../main.tex

\section{Experiments}

\subsection{Dataset and Experimental Setup}

We evaluate the temporal structures associated with game-theoretic assumptions against six widely used trajectory datasets: INTERACTION (abbreviated as INT. in tables)~\cite{zhan2019interaction}, highD~\cite{krajewski2018highd}, inD~\cite{bock2020ind}, rounD~\cite{krajewski2020round}, the Waymo Open Motion Dataset (WOMD)~\cite{ettinger2021large}, and nuPlan~\cite{caesar2021nuplan}.
These datasets provide substantial variation in data sources, road geometries, traffic densities, and driving environments, covering scenarios such as intersections, roundabouts, highways, and urban streets.
For all datasets, interaction events are extracted following the method in Sec.~\ref{sec:interaction-event-extraction}.
Raw trajectories are processed using Tactics2D-based parsers where supported, together with dataset-specific adapters for Waymo and nuPlan~\cite{li2024tactics2d}.
Dataset-specific statistics of the interaction events are summarized in Table~\ref{tab:interaction-event}.
The number of extracted events is lower than that reported by InterHub~\cite{jiang2025naturalistic}, as our definition additionally requires measurable behavioral change from at least one interacting vehicle.

% !TEX root = ../main.tex

\begin{table}[htb]
\centering
\caption{Extracted interaction events from different datasets.}\label{tab:interaction-event}
\begin{tabular}{l c c c c c}
\toprule[2pt]
\textbf{Dataset} & \textbf{\# Event} & \textbf{\# F} & \textbf{\# M} & \textbf{\# C} & \makecell{\textbf{Med.}\\\textbf{Duration (s)}}
\\ \midrule
\textbf{INT.} & 765 & 104 & 225 & 436 & 2.30 \\
\textbf{highD} & 1,387 & 1,324 & 63 & 0 & 3.48 \\
\textbf{inD} & 225 & 12 & 92 & 121 & 2.44 \\
\textbf{rounD} & 1,371 & 207 & 258 & 906 & 2.24 \\
\textbf{Waymo} & 5,565 & 3,078 & 1,996 & 491 & 2.70 \\
\textbf{nuPlan} & 2,566 & 1,588 & 662 & 316 & 2.65 \\
\bottomrule
\end{tabular}
\end{table}

\subsection{Temporal Organization of Vehicle Interactions}\label{sec:experiment-organization}

Using the complete sequence of detected behavioral changes described in Sec.~\ref{sec:temporal-characterization}, each event is classified as concurrent, sequential, or one-sided. Concurrent events have all behavioral changes within the 400~ms tolerance, whereas sequential events show a temporal ordering between the two vehicles.
The results are shown in Table~\ref{tab:temporal-organization}.
% !TEX root = ../main.tex

\begin{table}[htb]
\centering
\caption{Temporal organization of vehicle interactions across datasets and interaction types.}\label{tab:temporal-organization}
\begin{tabular}{l c c c c}
\toprule[2pt]
\textbf{Dataset} & \makecell{\textbf{Interaction}\\\textbf{type}} &
\makecell{\textbf{Concurrent}\\(\%)} & \makecell{\textbf{Sequential}\\(\%)} & \makecell{\textbf{One-sided}\\(\%)} \\
\midrule
\multirow{3}{*}{\textbf{INT.}} & F & 48.08 & 46.15 & 5.77 \\
    & M & 58.67 & 37.33 & 4.00 \\
    & C & 43.98 & 48.61 & 7.18 \\ \midrule
\multirow{2}{*}{\textbf{highD}} & F & 50.68 & 42.15 & 7.18 \\
    & M & 44.44 & 47.62 & 7.94 \\ \midrule
\multirow{3}{*}{\textbf{inD}}  & F & 45.45 & 54.55 & 0.00 \\
    & M & 54.26 & 44.68 & 1.06 \\
    & C & 46.15 & 50.43 & 3.42 \\ \midrule
\multirow{3}{*}{\textbf{rounD}} & F & 49.04 & 47.60 & 3.37 \\
    & M & 50.78 & 44.57 & 4.65 \\
    & C & 33.81 & 52.77 & 13.30 \\ \midrule
\multirow{3}{*}{\textbf{Waymo}} & F & 48.92 & 43.93 & 6.66 \\
    & M & 52.49 & 41.06 & 6.20 \\
    & C & 52.25 & 40.57 & 6.15 \\ \midrule
\multirow{3}{*}{\textbf{nuPlan}} & F & 52.09 & 40.96 & 6.83 \\
    & M & 57.51 & 33.69 & 8.80 \\
    & C & 55.41 & 34.08 & 10.51 \\
\bottomrule[2pt]
\end{tabular}
\end{table}

Concurrent events are common across datasets and interaction types, but they do not consistently account for the majority of interactions.
Across the reported cases, the concurrent proportion ranges from 33.81\% to 58.67\%.
Sequential events are also frequent, accounting for 33.69\%--54.55\% of events, and are more frequent than concurrent events in, for example, InD following and RounD conflicting interactions.
One-sided events are relatively rare, reaching at most 13.30\% for conflicting interactions in RounD.

\begin{figure}[htb]
    \centering
    \subfloat[Temporal organization]{\includegraphics[width=0.66\linewidth]{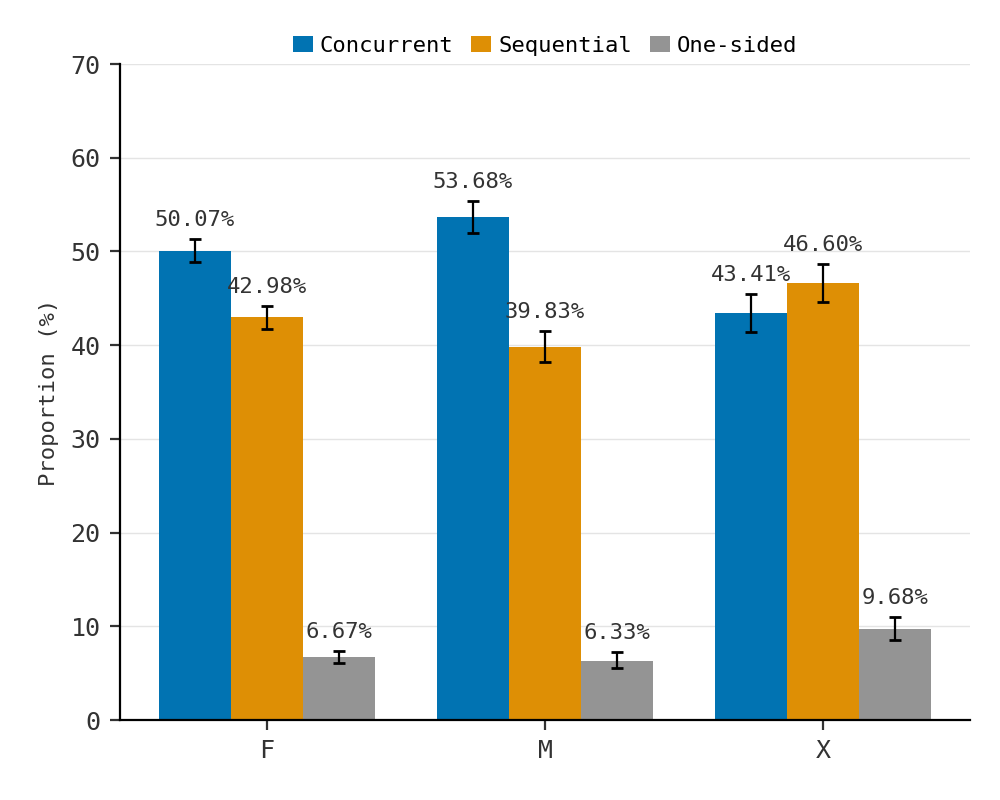}\label{fig:temporal-organization}}
    \hfill
    \subfloat[Adjusted odds ratios]{\includegraphics[width=0.33\linewidth]{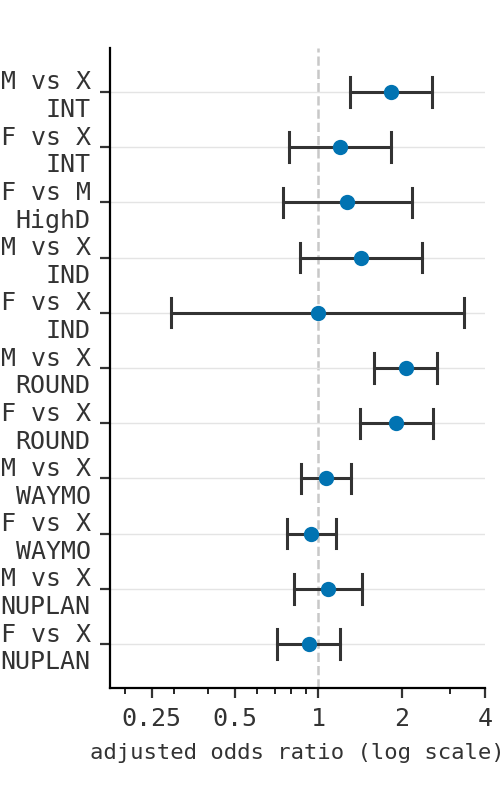}\label{fig:glmm}}
    \caption{Temporal organization of vehicle interactions and adjusted odds ratios across interaction types.}\label{fig:interaction-organization}
\end{figure}

Figure~\ref{fig:interaction-organization}~\subref{fig:temporal-organization} shows the distribution of temporal organization across interaction types.
The corresponding GLMM results are shown in Fig.~\ref{fig:interaction-organization}~\subref{fig:glmm}, where the reported odds ratios are adjusted for event duration, the number of vehicles, and scene-level heterogeneity.
Together, these results show that the temporal organization of vehicle interactions varies across interaction types, with both concurrent and sequential changes occurring frequently.

\subsection{Response Dynamics and Interaction Directionality}\label{sec:experiment-directionality}

For each temporally ordered pair, the vehicle that changes first is treated as the source and the other as the target.
A response is identified from the target's post-onset behavioral change, with the response lag measured from the source onset to the target response onset.

\begin{figure}[htb]
    \centering
    \subfloat[Response probability among assessable directed pairs]{\includegraphics[width=0.49\linewidth]{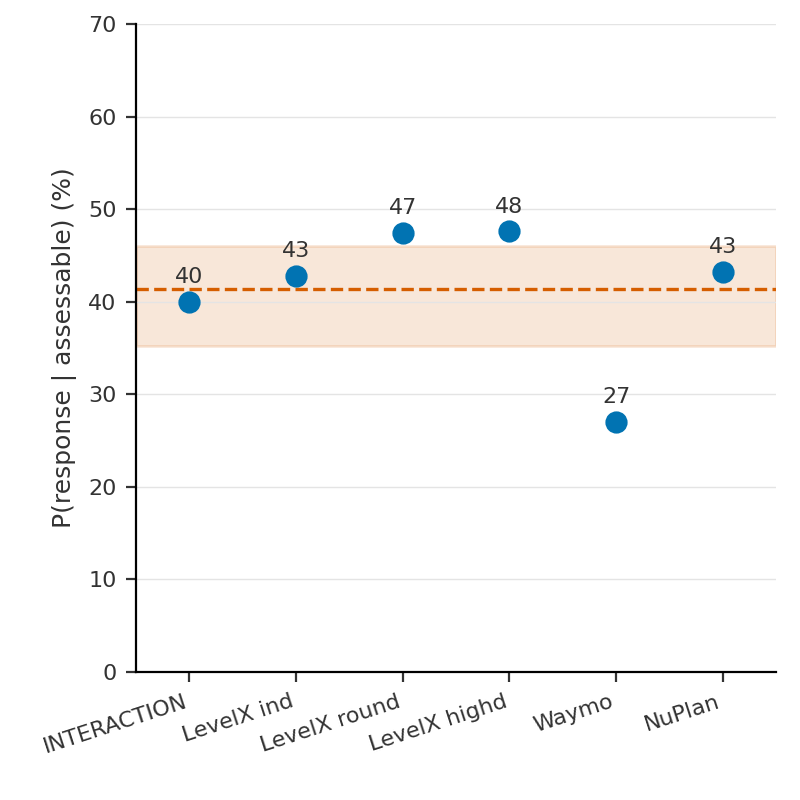}\label{fig:response-probability}}
    \hfill
    \subfloat[Dataset-averaged empirical cumulative distribution of response lags]{\includegraphics[width=0.49\linewidth]{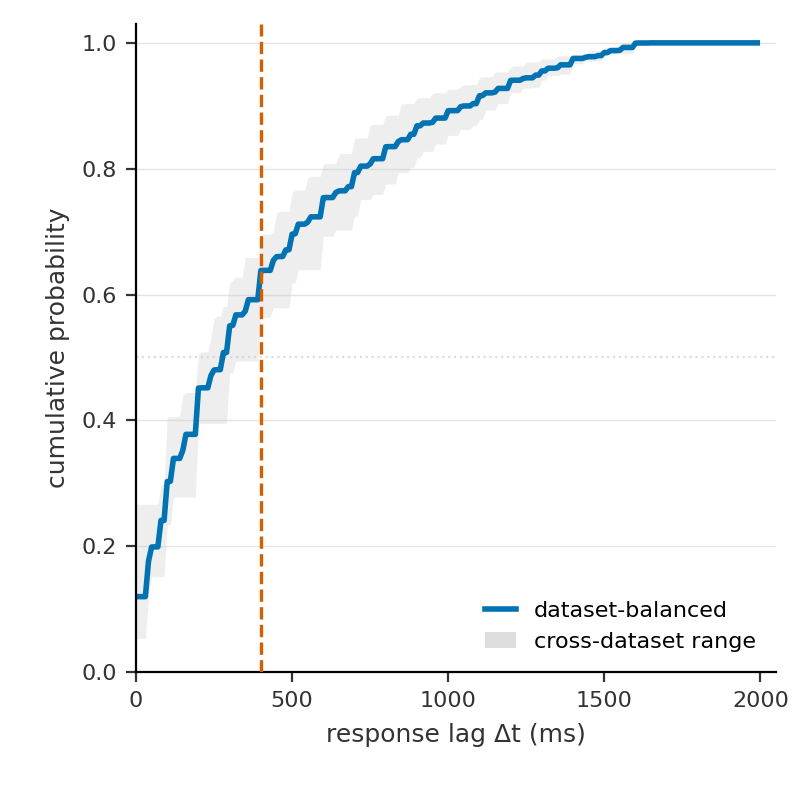}\label{fig:ECDF}}
    \caption{Response dynamics across datasets.}\label{fig:response-dynamics}
\end{figure}

Figure~\ref{fig:response-dynamics}\subref{fig:response-probability} shows the response probability among assessable directed pairs, defined as directions classified as either response or preactive.
Directions with no valid source onset, insufficient pre-onset baseline frames, no response, or right-censoring are excluded.
The response probability is 40.0\% in INTERACTION, 42.8\% in InD, 47.5\% in RounD, 47.6\% in HighD, 27.1\% in Waymo, and 43.2\% in nuPlan.
With equal weighting across datasets, the mean response probability is 41.4\% (95\% CI: 35.2--46.0\%).

Figure~\ref{fig:response-dynamics}~\subref{fig:ECDF} shows the dataset-averaged empirical cumulative distribution function (ECDF) of response-lag.
About 64\% of detected responses occur within the 400~ms concurrent tolerance, ranging from 56.4\% to 69.5\% across datasets.

These results show that temporal precedence does not necessarily lead to a measurable behavioral response: only about 41\% of assessable directed pairs exhibit a response. When a response does occur, however, it is often observed within the 400~ms temporal tolerance.

\subsection{Ordering Stability}

We next examine whether a clear temporal ordering, once established, is maintained throughout the interaction.
Restricting attention to the sequential events identified in Sec.~\ref{sec:experiment-organization}, we ask whether the ordering is stable or alternating.
The results are reported in Table~\ref{tab:role-stability}.

% !TEX root = ../main.tex

% Ordering stability (conditional on sequential) by dataset and interaction type.
% Values are % of sequential interactions.
\begin{table}[htb]
\centering
\caption{}\label{tab:role-stability}
\setlength{\tabcolsep}{4pt}
\begin{tabular}{l c c c}
\toprule[2pt]
\textbf{Dataset} & \textbf{Interaction type} & \textbf{Stable (\%)} & \textbf{Alternating (\%)} \\
\midrule
\multirow{3}{*}{\textbf{INT.}} & F & 87.50 & 12.50 \\
    & M & 86.90 & 13.10 \\
    & C & 87.14 & 12.86 \\ \midrule
\multirow{2}{*}{\textbf{highD}} & F & 98.39 & 1.61 \\
    & M & 96.67 & 3.33 \\ \midrule
\multirow{3}{*}{\textbf{inD}}  & F & 100.00 & 0.00 \\
    & M & 80.95 & 19.05 \\
    & C & 84.75 & 15.25 \\ \midrule
\multirow{3}{*}{\textbf{rounD}} & F & 91.92 & 8.08 \\
    & M & 95.65 & 4.35 \\
    & C & 95.80 & 4.20 \\ \midrule
\multirow{3}{*}{\textbf{Waymo}} & F & 80.46 & 19.54 \\
    & M & 80.61 & 19.39 \\
    & C & 82.83 & 17.17 \\ \midrule
\multirow{3}{*}{\textbf{nuPlan}} & F & 83.80 & 16.20 \\
    & M & 91.89 & 8.11 \\
    & C & 88.79 & 11.21 \\
\bottomrule[2pt]
\end{tabular}
\end{table}

Stable ordering accounts for 80.46\%--100.00\% of sequential events across datasets and relation types, with alternating ordering accounting for the remainder.
Stable ordering reaches 100.00\% for following interactions in InD and exceeds 95\% for following and merging interactions in HighD and for merging and conflicting interactions in RounD.
Alternating ordering is most frequent in Waymo, accounting for 19.54\%, 19.39\%, and 17.17\% of sequential following, merging, and conflicting interactions, respectively.
Overall, temporal ordering is predominantly stable once a sequential interaction is established, while role alternation remains a minority pattern.

\subsection{Implications for Game-Theoretic Interaction Models}

The empirical findings provide the following implications for the game-theoretic assumptions introduced in Sec.~\ref{sec:preliminary}.

\subsubsection{Simultaneous-move game assumption}

Concurrent behavioral changes are common across datasets and interaction types, indicating that the simultaneous-move structure underlying Nash formulations represents a meaningful class of vehicle interactions.

\subsubsection{Sequential-move game assumption}

Sequential behavioral changes are also prevalent, with their proportion sometimes exceeding that of concurrent interactions.
This finding shows that sequential-move formulations are likewise relevant to observed vehicle interactions.

\subsubsection{Leader--follower structure}

Among sequential interactions, stable ordering is substantially more common than alternating ordering across datasets.
This suggests that persistent asymmetric ordering is a meaningful structure within sequential interactions, supporting the use of leader--follower formulations.

Overall, the results suggest a \textbf{mixture of interaction structures} rather than a single universal temporal organization.
Concurrent, sequential, and stably ordered behavioral patterns are all observed across different interaction contexts and datasets, indicating that different game-theoretic formulations provide complementary descriptions of real-world vehicle interactions.

% !TEX root = ../main.tex

\section{Conclusion}

This work presents a measurement framework for examining the temporal interaction structures associated with game-theoretic formulations using real-world vehicle trajectories.
Across six real-world trajectory datasets, vehicle interactions show substantial variation in temporal organization.
Concurrent and sequential behavioral changes both occur across following, merging, and conflicting scenarios, while sequential interactions are more often characterized by stable than alternating ordering.
These results indicate that real-world driving does not follow a single temporal interaction pattern.
Concurrent behavioral changes support simultaneous-move temporal formulations, whereas sequential interactions with stable ordering support leader--follower temporal formulations.
The same geometric relation can exhibit different temporal organizations across datasets, suggesting that interaction geometry alone is not sufficient to determine the appropriate interaction structure.
The response analysis also shows that temporal precedence does not necessarily imply a measurable behavioral response.
Temporal ordering and behavioral dependence should therefore be considered separately when interpreting sequential interactions.
This evidence concerns the temporal organization of observable behavior. It does not by itself establish vehicles' decision-making processes, information sets, utility functions, strategies, or equilibrium mechanisms.
Future interaction models should account for such variation in temporal organization and response behavior rather than assuming a single interaction mechanism.

\bibliographystyle{IEEEtran}
\bibliography{main}

% \normalsize
\begin{IEEEbiography}[{\includegraphics[width=1in,height=1.25in,clip,keepaspectratio]{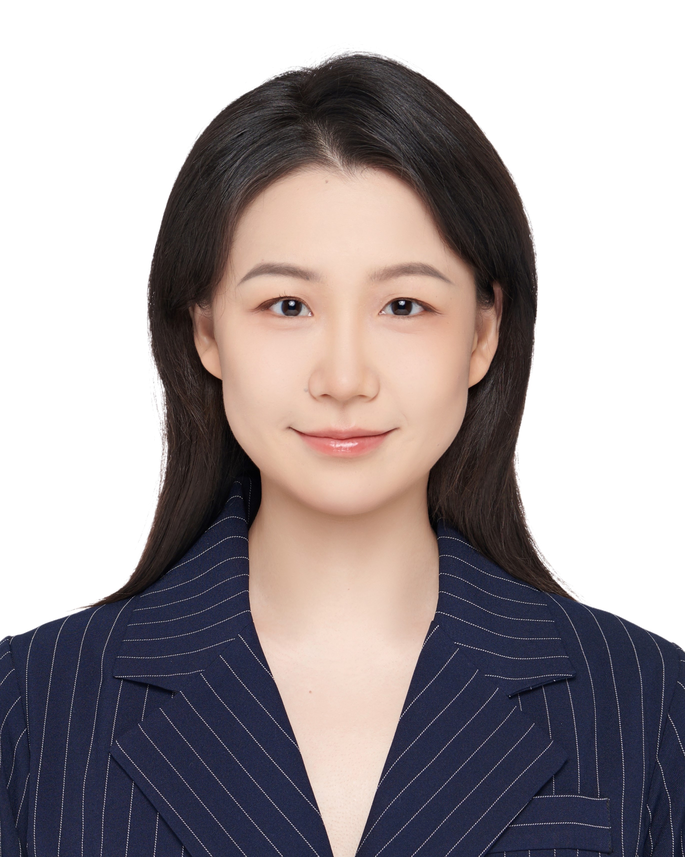}}]{Yueyuan LI}
    received a Bachelor's degree in Electrical and Computer Engineering from the University of Michigan-Shanghai Jiao Tong University Joint Institute in 2020. She is pursuing a Ph.D. degree in Control Science and Engineering from Shanghai Jiao Tong University. Her main fields of interest are the security of the autonomous driving system and driving decision-making. Her research activities include reinforcement learning, behavior modeling, and simulation.
\vspace{-20pt}
\end{IEEEbiography}

\begin{IEEEbiography}[{\includegraphics[width=1in,height=1.25in,clip,keepaspectratio]{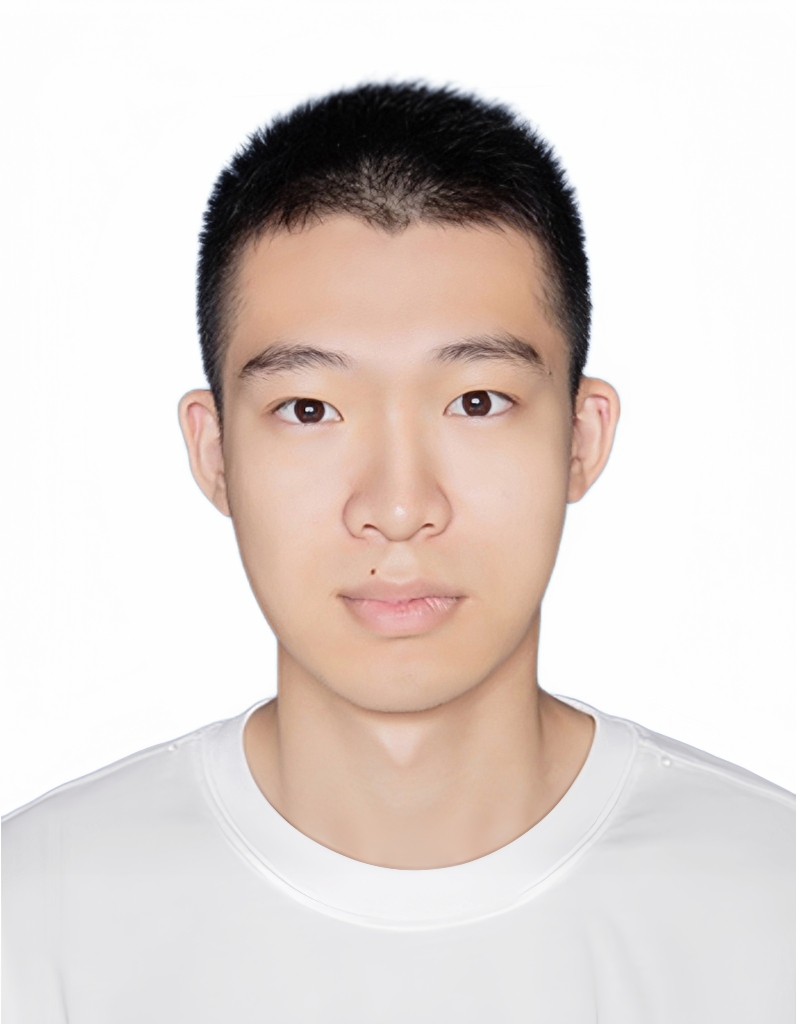}}]{Rongcheng NIE}
    is currently pursuing a Bachelor's degree in Automation with the School of Airspace Science and Engineering, Shandong University. His research interests include autonomous driving, robotics and embodied intelligence.
\end{IEEEbiography}

\begin{IEEEbiography}
[{\includegraphics[width=1in,height=1.25in,clip,keepaspectratio]{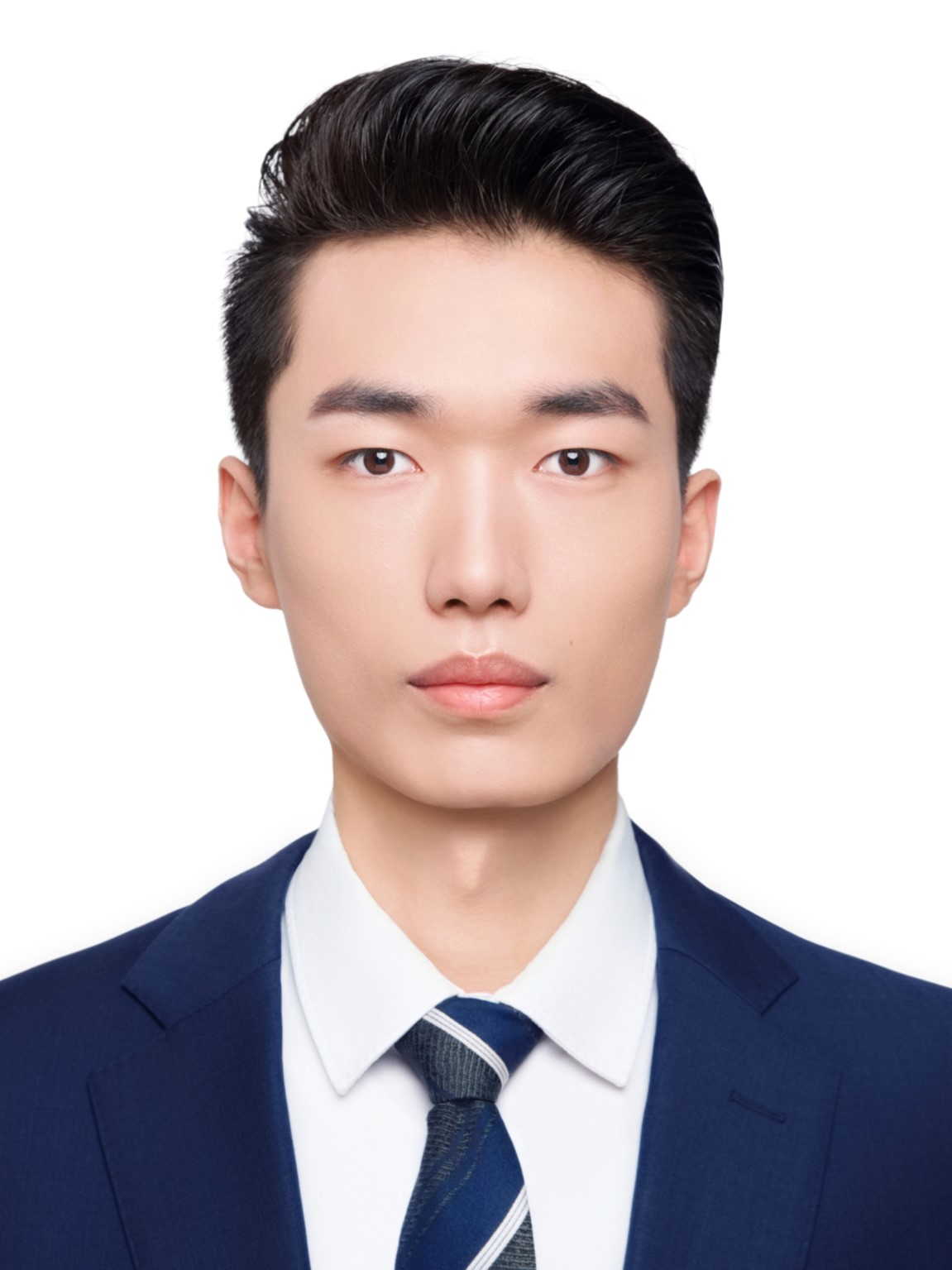}}]{Weijie XI}
    received a Bachelor’s degree in engineering from Jiangnan University, Jiangsu, China, in 2026. He is working towards a Ph.D. degree in Control Science and Engineering from Shanghai Jiao Tong University. His research interests include learning-based planning and control for intelligent vehicles and mobile robots.
\vspace{-20pt}
\end{IEEEbiography}

\begin{IEEEbiography}
[{\includegraphics[width=1in,height=1.25in,clip,keepaspectratio]{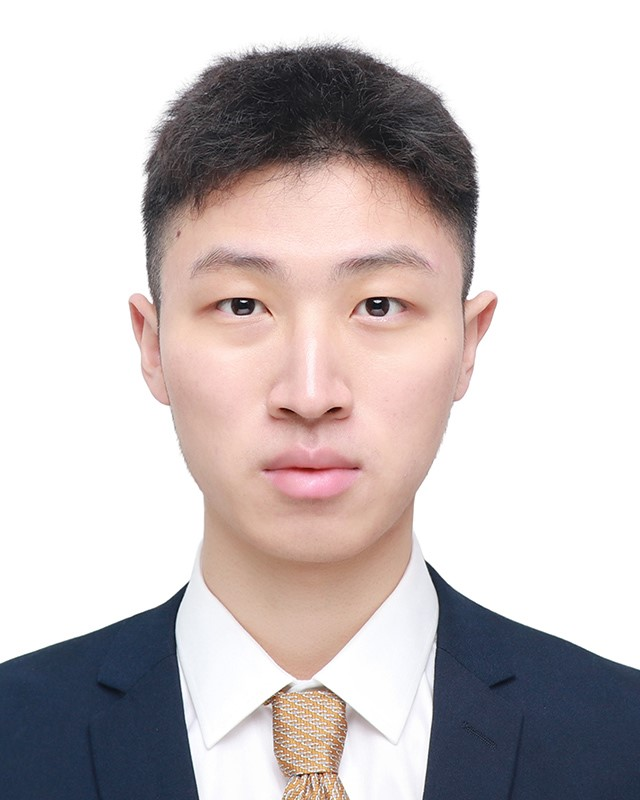}}]{Mingyang JIANG}
    received a Bachelor's degree in engineering from Shanghai Jiao Tong University in 2023, and a Master's degree in Control Science and Engineering from Shanghai Jiao Tong University in 2026. His main research interests are end-to-end planning, driving decision-making, and reinforcement learning for autonomous vehicles.
\vspace{-20pt}
\end{IEEEbiography}

\begin{IEEEbiography}[{\includegraphics[width=1in,height=1.25in,clip,keepaspectratio]{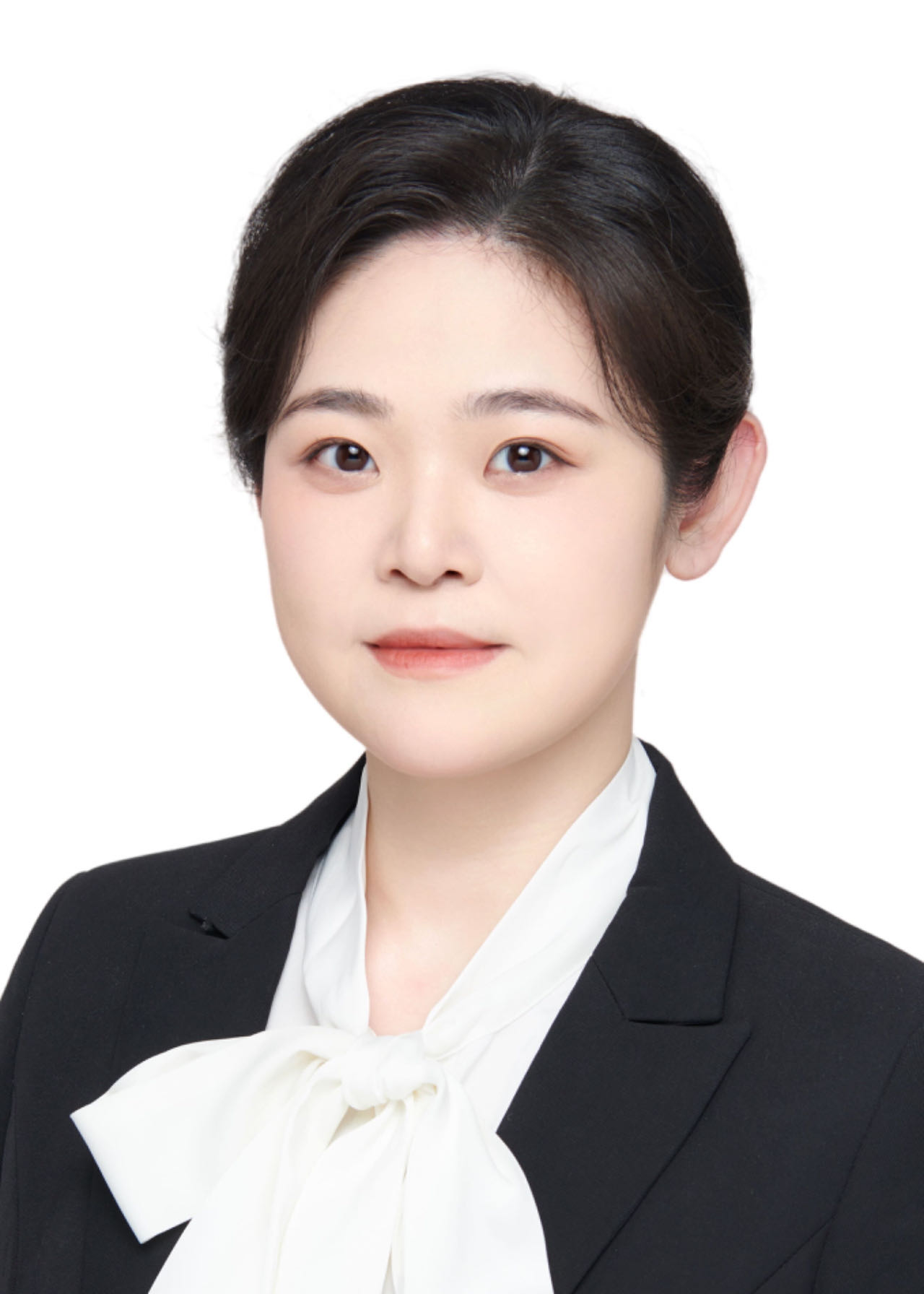}}]{Songan ZHANG}
    received B.S. and M.S. degrees in automotive engineering from Tsinghua University in 2013 and 2016, respectively. Then, she went to the University of Michigan, Ann Arbor, and received a Ph.D. in mechanical engineering in 2021. After graduation, she worked as a research scientist on the Robotics Research Team at Ford Motor Company. Presently, she is an assistant professor at the Global Institute of Future Technology (GIFT) at Shanghai Jiao Tong University. Her research interests include accelerated evaluation of autonomous vehicles, model-based reinforcement learning, and meta-reinforcement learning for autonomous vehicle decision-making.
\vspace{-20pt}
\end{IEEEbiography}

\begin{IEEEbiography}[{\includegraphics[width=1in,height=1.25in,clip,keepaspectratio]{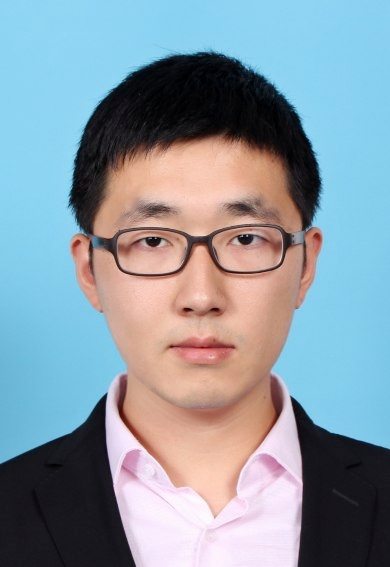}}]{Hanyang ZHUANG}
    received the Ph.D. degree from Shanghai Jiao Tong University, Shanghai, China, in 2018. He has worked as a postdoctoral researcher at Shanghai Jiao Tong University from 2020 to 2022. He is currently an assistant research professor at Shanghai Jiao Tong University implementing research works related to intelligent vehicles. His research interest is in autonomous driving and cooperative driving systems.
\vspace{-20pt}
\end{IEEEbiography}

\begin{IEEEbiography}[{\includegraphics[width=1in,height=1.25in,clip,keepaspectratio]{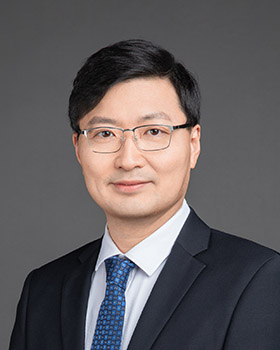}}]{Ming YANG}
    received his Master’s and Ph.D. degrees from Tsinghua University, Beijing, China, in 1999 and 2003, respectively. Presently, he holds the position of Distinguished Professor at Shanghai Jiao Tong University, also serving as the Director of the Innovation Center of Intelligent Connected Vehicles. Dr. Yang has been engaged in the research of intelligent vehicles for more than 25 years.
\vspace{-20pt}
\end{IEEEbiography}

\vspace{30pt}

\end{document}